\documentclass[conference]{IEEEtran}
\IEEEoverridecommandlockouts
\usepackage{cite}
\usepackage{amsmath,amssymb,amsfonts}
\usepackage[draft]{minted}

\usepackage{caption}
\usepackage{lineno}
\usepackage{algorithm}
\usepackage{algpseudocode}
\usepackage{graphicx}
\usepackage{textcomp}
\usepackage{xcolor}
\usepackage{booktabs}
\usepackage{subfigure}
\usepackage{svg}
\usepackage{dblfloatfix}
\def\BibTeX{{\rm B\kern-.05em{\sc i\kern-.025em b}\kern-.08em
    T\kern-.1667em\lower.7ex\hbox{E}\kern-.125emX}}
\begin{document}

\title{Enhancing the Convergence of Federated Learning Aggregation Strategies with Limited Data
}

\author{\IEEEauthorblockN{1\textsuperscript{st} Judith Sáinz-Pardo Díaz}
\IEEEauthorblockA{\textit{Instituto de Física de Cantabria (IFCA), CSIC-UC}\\
Avda. los Castros s/n. 39005 - Santander (Spain) \\
sainzpardo@ifca.unican.es}
\and
\IEEEauthorblockN{2\textsuperscript{nd} Álvaro López García}
\IEEEauthorblockA{\textit{Instituto de Física de Cantabria (IFCA), CSIC-UC}\\
Avda. los Castros s/n. 39005 - Santander (Spain) \\
aloga@ifca.unican.es}
}

\maketitle

\begin{abstract}
The development of deep learning techniques is a leading field applied to cases in which medical data is used, particularly in cases of image diagnosis. This type of data has privacy and legal restrictions that in many cases prevent it from being processed from central servers. However, in this area collaboration between different research centers, in order to create models as robust as possible, trained with the largest quantity and diversity of data available, is a critical point to be taken into account. In this sense, the application of privacy aware distributed architectures, such as federated learning arises. When applying this type of architecture, the server aggregates the different local models trained with the data of each data owner to build a global model. This point is critical and therefore it is fundamental to analyze different ways of aggregation according to the use case, taking into account the distribution of the clients, the characteristics of the model, etc. In this paper we propose a novel aggregation strategy and we apply it to a use case of cerebral magnetic resonance image classification. In this use case the aggregation function proposed manages to improve the convergence obtained over the rounds of the federated learning process in relation to different aggregation strategies classically implemented and applied.
\end{abstract}


\section{Introduction}

Artificial intelligence (AI) applications are transforming the field of healthcare in several areas, from image-based classification, to data-driven diagnostic support and improvement, drug discovery, early detection and clinical outcome prediction of pathologies and biomarkers discovery among others \cite{yu2018artificial, bohr2020artificial}. One of the major problems that arises when applying a AI based models, namely machine and deep learning (ML/DL) to data from the clinical sector concerns the privacy restrictions of the data. In Europe, the General Data Protection Regulation (GDPR) \cite{gdpr} establishes in its article 9 a general regime for the processing of clinical data, which is expressly prohibited unless there are different circumstances, such as that it is necessary to protect the vital interest of the data subject, that there is an essential public interest reason, that there is the explicit consent of the interested party or that the data is manifestly made public by the individual concerned are some of them. In addition, the new European regulation for artificial intelligence, the AI Act \cite{AIAct}, seeks to define European standards for the development of AI-based solutions and products in the European Union. This legislation seeks to establish different regulations based on the different levels of risks defined, which will have a direct impact on the development of this type of solutions in the field of healthcare where highly sensitive data is handled.  

Different privacy preserving techniques can be employed when dealing with structured medical data, from anonymization and psedudonymization to the application of differential privacy techniques or even the training of models using homomorphic encryption procedures. However, this task is more complicated when the available data are images, and even more so when we have different data providers that, due to privacy issues of patient data, cannot download the data from their own servers. 

In this sense, a widely used alternative is to train AI/ML/DL models on decentralized data, i.e., without the data owners having to take their data from their own location, even for training. In this regard, the concept of federated learning (FL) arises \cite{mcmahan2017communication}, where there is an architecture consisting of multiple clients (each data owner) and a central server. The clients train the same model and the server only receives its parameters or weights, and aggregates them to build a global model. Details about this architecture will be discussed in the next section.

As mentioned above, a key step is the construction of the global model by aggregating the local models from all the data owners, for which different aggregation strategies can be used. In this work, we propose and introduce a novel aggregation strategy named \textit{FedAvgOpt} and we analyze its applicability to a use case of classification of brain  Magnetic Resonance Imaging (MRI). Specifically, we perform the experiments simulating few data available for training.

In this sense, in this work we will perform an implementation of a federated learning architecture for the classification of brain magnetic resonance imaging (MRI) according to the pathology presented by the patient. Regarding the diagnosis of brain tumors, it is important to start by highlighting that they compose between 85$\%$ and 90$\%$ of the primary Central Nervous System (CNS) tumors \cite{colopi2023impact} and as stated in \cite{watanabe1992magnetic}, MRI can be used for predicting the grade of cerebral gliomas. Gliomas are the most common brain tumors in adults, accounting for 30$\%$ of all brain and central nervous system tumors \cite{schneider2010gliomas} and 80$\%$ of all malignant brain tumors \cite{indira2024extracellular}. Misclassification of these pathologies can lead to prescription of wrong treatments that could cause worsening of the condition, so applying AI for such classification can help practitioners and support proper diagnosis. 

Thus, in this work we are going to take a dataset available in open source and the objective is to classify brain MRI images in 4 categories: glioma, meningioma, pituitary tumor and no tumor. For this purpose we will implement a FL architecture, simulating the case in which the data could not be centralized for various technical, legal and privacy reasons. Therefore, the data will be distributed among 4 disjoint clients. Different models will be used to perform the classification and we will simulate a use case in which limited data are available for training.

This paper is structured as follows: in Section~\ref{sec:fl} the concept of federated learning is presented together with the aggregation strategies that will be compared in this study. Section~\ref{sec:sota} analyzes the state of the art both in relation to FL applications to medical data and to aggregation strategies and benchmarks. Section~\ref{sec:fedavgopt} introduces the novel aggregation strategy proposed in this work. Section~\ref{sec:methodology} presents the methodology applied, detailing the problem studied, the data used and the DL models developed. The results obtained with each base model and with each aggregation function are detailed and compared in Section~\ref{sec:results}. Finally, Section~\ref{sec:conclusions} contains the conclusions and gives some lines of future work.

\section{Federated Learning: aggregation strategies}\label{sec:fl}

Classically, the federated learning architecture (note that in this work we always mean horizontal FL when referring to federated learning), is composed of a central server and multiple clients (data owners). This architecture can be seen as a concrete case of distributed machine learning, where the concept of data parallelism is applied, since the data are distributed among different workers (in this case one worker for each data owner).

\begin{figure}[ht]
    \centering
    \includegraphics[width=\linewidth]{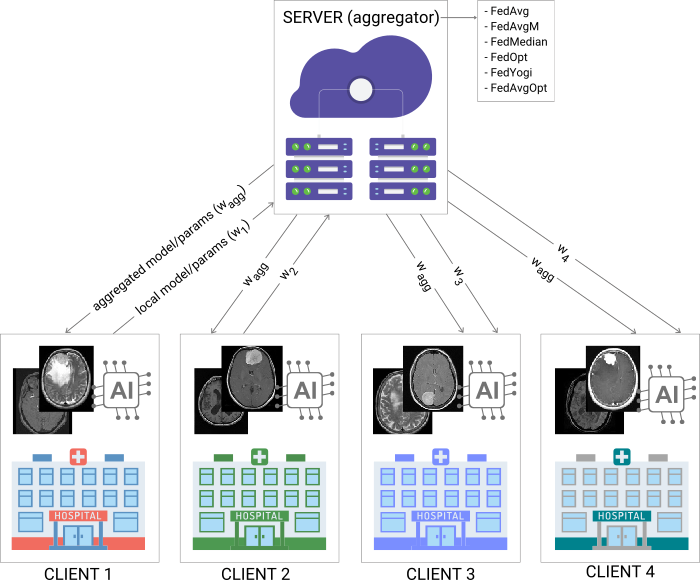}
    \caption{Federated learning architecture applied to the current use case.}
    \label{fig:fl-mri}
\end{figure}

Specifically, the structure that has been followed in this work is the following (summarized in a schematic form in Figure~\ref{fig:fl-mri} for the present use case under study) \cite{sainzpardo2023study}. 

\begin{itemize}
    \item[1.] A model to be trained by the clients is consensually defined. All clients start training with the same model. The clients select the aggregation function they would like to use, as well as the number of rounds to be trained and the metric to be computed in an aggregated form. 
    \item[2.] Each data owner trains the model locally with their data.
    \item[3.] The parameters defining the resulting model, the gradients (in case of a neural network), the model itself or the model updates are sent to the central server by each client.
    \item[4.] The server aggregates the models received by the participating clients by applying the pre-established aggregation function.
    \item[5.] The aggregated global model created by the server is distributed to each client, to re-train it again with its data and re-send it to the server. This process is repeated for a pre-determined number of rounds.
\end{itemize}

The stage in which the aggregation is performed is critical, since it is the point at which the global model that will be tested by the different parties involved is built. In this work we are interested in comparing different classical aggregation strategies in relation to a new proposed solution, in order to analyze the convergence and performance of each of them in a medical imaging case using different base model. Below are some aggregation strategies that do not require additional information to be added from the client side, and that will be compared in this work together with the new proposed strategy. 

Note that for testing this functions and implementing the whole federated learning architecture the open source Python library \textit{Flower} will be used \cite{beutel2020flower}.

\subsection{Federated Average (FedAvg)}

The Federated Averaging strategy (\textit{FedAvg}) is the most widely used aggregation function in the literature for its simplicity and effectiveness. The idea is to simple average the local models in a weighted way taking into account the number of data available for each client. It was proposed by B. McMahan et al. in 2017 \cite{mcmahan2017communication} when introducing the FL architecture.

\subsection{Federated Average with Momentum (FedAvgM)}

This strategy is implemented in \cite{fedavgm}, and it aims to incorporate the idea of momentum at the central server side during the aggregation process to improve the stability and speed of convergence of the overall model. 

Then, for getting the aggregated parameters using \textit{FedAvgM} we have to calculate the momentum vector in round $r$ $v^{r}$ (initialized as $v_{0}=0$), in such a way that $v^{r+1}=\beta v^{r} + \Delta w^{r+1}$. Then, we will calculated the weights in round $r+1$ as $w^{r+1}=w^{r}-v^{r+1}$, being $\beta$ the momentum. %(note that with FedAvg we will have $w^{r+1}=w^{r}-\Delta w^{r+1}$) Additionally, we can add the term $\mu$ as server learning rate in both cases multiplying it by $v^{r+1}$ or $\Delta w^{r+1}$ in each case.
Additionally, we can add the term $\mu$ as server learning rate multiplying it by $v^{r+1}$.

\subsection{Federated Median (FedMedian)}

The Federated Median strategy (\textit{FedMedian}) was proposed in \cite{fedmedian}. Specifically, the proposed algorithm for aggregating using the median takes into account byzantine machines so that only in the case of normal worker machines each such client computes the local gradient. Be $w^{r}$ the weights in round $r$, the central server will be in charge of getting the median of the local gradients in each round $r$ ($g_{median}(w^{r})$). Then, once such aggregated gradient is calculated, the weights are updated for round $r+1$: $w^{r+1}=w^{r}-\mu g_{median}(w^{r})$ \cite{fedmedian}, being $\mu$ the step-size or learning rate.      

\subsection{Adaptive Federated Optimization (FedOpt and FedYogi)}

Other aggregation function available in the literature that we can highlight is Adaptive Federated Optimization (\textit{FedOpt}) \cite{reddi2020adaptive}. The pseudocode for implementing the \textit{FedOpt} strategy is shown in Algorithm 1 from \cite{reddi2020adaptive}, and it aims to improve \textit{FedAvg} by using gradient-based optimizers with given learning rates that must be customized both from the client and server side. %Then, be $w^{r}$ the global weights at round $r$, $w_{i}^{r}$ the weights for client $i$, then we can define $\Delta^{r}_{i} = w_{i}^{r} - w^{r}$ and $\Delta^{r} = \frac{1}{n}\sum_{i=1}^{n}\Delta^{r}_{i}$. From the server side, we will use an optimizer for calculating $w^{r+1}$ given $w^{r}$, $\Delta^{r}$ and the learning rate. 

Note that \textit{FedOpt} allows the use of different adaptive optimizers in the server side, such us \textit{Adam}, \textit{Adagrad} or \textit{Yogi} (\textit{FedAdam}, \textit{FedAdagrad} and \textit{FedYogi}). The adaptive methods are used on the server side while SGD is used from the client side \cite{reddi2020adaptive}. More details are given in the pseudocode of Algorithm 2 from \cite{reddi2020adaptive}.

%These strategies require additional parameters, such as the server-side and client-side learning rate, $\beta_{1}$ and $\beta_{2}$ as momentum and second momentum parameters respectively and $\tau$ for controlling the degree of adaptability of the algorithm. According to its implementation in Flower, some initial global parameters for the model need to be introduced when using these functions. The use of these global initial parameters provides a consistent starting point for all clients.

Finally, this paper will compare the performance obtained for a brain MRI classification use case with limited data for training applying FL with the following aggregation functions: \textit{FedAvg}, \textit{FedAvgM}, \textit{FedMedian}, \textit{FedOpt}, \textit{FedYogi} and the proposed strategy (\textit{FedAvgOpt}).

\section{State of the Art}\label{sec:sota}

The previous section has reviewed various classic aggregation strategies that will be initially compared in this paper along with the novel proposed one. Additionally, other aggregation functions that involve some changes also from the client side have been defined in the literature. For example, \textit{FedProx} aims ``to tackle heterogeneity in federated networks'', being a generalization and re-parametrization of the \textit{FedAvg} strategy \cite{li2020federated}. In \cite{QI2024272} different aggregation functions are reviewed, such as \textit{MOON} \cite{li2021model}, which aims to minimize the discrepancy between the client's local models and the global one by incorporating a contrastive model loss as a regularization term. The Scaffold strategy is also reviewed, which tries to address the client-drift problem resulting from data heterogeneity \cite{karimireddy2020scaffold}, and \textit{FedNova} \cite{wang2020tackling}, which seeks to enhance the model aggregation of the \textit{FedAvg} method to address non i.i.d conditions.  %Additionally, other aggregation functions that involve some changes also from the client side have been defined in the literature. For example, \textit{FedProx} aims ``to tackle heterogeneity in federated networks'', being a generalization and re-parametrization of the \textit{FedAvg} strategy \cite{li2020federated}. This technique is proven to be more stable and accurate than \textit{FedAvg} in highly heterogeneous settings \cite{li2020federated}. In addition, \textit{FedProx} by its very definition involves modifying the loss function locally on the side of each client by adding the regularization with the proximal $\mu$ term. This term can be introduced from the server side and received by each client, so that they all use the same one. However, the client-side code has to be adapted appropriately in each case concerning the local loss function. 

Regarding the application of FL to cases in the field of brain tumor classification via MRI, in \cite{islam2023effectiveness} the authors analyze brain tumor images obtained from \cite{pernet2016neuroimaging} and apply a FL architecture to a binary classification problem (tumor or not) using different CNN architectures (VGG19, InceptionV3, DenseNet121 and an ensemble). Specifically, an accuracy greater than 90$\%$ is obtained for the binary classification. In \cite{zhou2024distributed} five clients are analyzed in a FL architecture to classify MRI images for tumor classification, using two CNNs, EfficientNetB0 with an accuracy of 80.17$\%$ and ResNet-50 with an accuracy of 65.3$\%$. The authors use the same data as in the current study (then it's a classification problem with four categories) although with the images scaled with 512x512 resolution and using four images less. More recently, in 2024 \cite{DALMAZ2024103121} a novel personalized federated learning method (pFLSynth) for multi-contrast MRI synthesis is proposed. In that work, other methods are compared, such as FedMRI, proposed in \cite{feng2022specificity} as a specificity-preserving FL algorithm for MRI image reconstruction.

Concerning FL applied to medical imaging, \cite{sohan2023systematic} presents a systematic review of 17 papers covering different types of medical images, from lung X-ray images, brain MRI to desmoscopy and retinal images among others. In such papers ML/DL models and FL algorithms are implemented.

\section{Federated Average Optimized: FedAvgOpt}\label{sec:fedavgopt}

The idea of the aggregation strategy proposed in this work, named \textit{FedAvgOpt} (as it aims to be an optimized version of \textit{FedAvg}), is to seek that the aggregated weights are at minimum distance from the individual weights calculated for each of the clients. Then, be $n_{i}$ the number of data of each client $i$, and $w_{i}^{r}$ the weights associated with the model trained with data from client $i$ $\forall i \in \{1,\hdots,n\}$ after round $r$. Be $n$ the total number of clients participating in the FL training. We define $w_{FedAvg}^{r}$ as the aggregated weights obtained using \textit{FedAvg} strategy in round $r$. The aggregated weights obtained using \textit{FedAvgOpt} ($w_{FedAvgOpt}^{r}$) after training in each client in round $r$ are defined as follows:

\begin{equation*}
    w_{FedAvgOpt}^{r} = \frac{1}{\sum_{i=1}^{n}n_{i}}\sum_{i=1}^{n}w_{i}^{r}n_{i}\alpha_{i}^{r},
\end{equation*}

In this sense, it is necessary to solve the following nonlinear optimization problem to obtain the value of $\alpha$ for each round $r$ and subsequently calculate $w_{FedAvgOpt}^{r}$.

\begin{equation*}
(P)\left\{ \begin{array}{l} 
    \min f(\alpha)\\
    \mbox{\textit{subject to }} \alpha \in \mathbb{R}^{n}
    \end{array}
\right.,
\end{equation*}

with the function $f(x)$ continuous in the domain in which it is defined by being the finite sum of continuous functions and defined below:

\begin{equation}
    f(x) =\sum_{j=1}^{n}\frac{||\frac{1}{\sum_{i=1}^{n}n_{i}}\sum_{i=1}^{n}w_{i}n_{i}x_{i} - w_{j}||_{2}}{||\frac{1}{\sum_{i=1}^{n}n_{i}}\sum_{i=1}^{n}w_{i}n_{i}x_{i} + w_{j}||_{2}},
\end{equation}

or, analogously:

\begin{equation}\label{eq:fx}
    f(x) = \sum_{j=1}^{n}\frac{||w_{FedAvg}x - w_{j}||_{2}}{||w_{FedAvg}x + w_{j}||_{2}}.
\end{equation}

Note than in $f(x)$, $w_{j}$ correspond to the weights calculated for client $j$ $\forall j\in \{1,\hdots,n\}$ in the current round $r$, and $w_{FedAvg}$ are the aggregated weights obtained using \textit{FedAvg} for such round (calculated after sending to the server $\{w_{j}: \forall j\in \{1,\hdots,n\} \}$).

In order to solve $(P)$, the Nelder-Mead method can be used, which is a simplex algorithm that allows to minimize a function in a multidimensional space (non-linear simplex). This algorithm is specially designed for unconstrained optimization, and it was introduced in 1965 by J. Nelder and R. Mead. This method requires an initial point $x_0$ for starting the optimization, and in this case we take $x_{0} = \boldsymbol{1}_{n}=(1,\hdots,1)$ with the $n$ the number of clients. Note that if $\alpha^{r}=1_{n}$ for a certain round $r$, then $w_{FedAvgOpt}^{r}=w_{FedAvg}^{r}$.

\begin{algorithm}[H]
	\caption{FedAvgOpt}\label{alg:fedavgopt}
    \textbf{\textit{Input}}: $N_{r}$ (number of rounds), $\mathcal{C}$ (set of clients), $n$ (array with the number of data in each client), $\mathcal{M}$ (model to be trained), $f$ (function to be optimized)
	\begin{algorithmic}[1]
        \State Initialize the weights $w^{0}$ for the model $\mathcal{M}$
        \State Initialize $x_{0}=\boldsymbol{1}_{|\mathcal{C}|}$
			\For{$r \in \{1,\hdots,N_r\}$}
                \State Initialize an empty array $w_{c}^{r}$ of length $|\mathcal{C}|$
                \For{each client $i \in \mathcal{C}$}
                \State Receive $w^{r-1}$ and build the model $\mathcal{M}(w^{r})$ 
			    \State Get $w_{i}^{r}$ after training $\mathcal{M}(w^{r})$  in client $i$
                \State Save $w_{i}^{r}$ and add it to $w_{c}^{r}$
				\EndFor
			\State $w_{FedAvg}^{r}$ $\gets$ $\frac{1}{\sum_{i=1}^{n}n_{i}}\sum_{i=1}^{n} n_{i} w_{i}^{r}$
            \State $\alpha^{r}$ $\gets$ \textit{minimize($f, x_{0}, w_{FedAvg}^{r}, w_{c}^{r}$)}
            \State $w_{FedAvgOpt}^{r}$ $\gets$ $\frac{1}{\sum_{i=1}^{n}n_{i}}\sum_{i=1}^{n} n_{i} w_{i}^{r}\alpha_{i}^{r}$
            \State \State $w^{r}$ $\gets$ $w_{FedAvgOpt}^{r}$
            \EndFor
			\end{algorithmic}
		\end{algorithm}

The pseudocode for implementing \textit{FedAvgOpt} is given in Algorithm~\ref{alg:fedavgopt}, using as input the set of clients $\mathcal{C}$ (each client with $n_{i}$ data $\forall i \in \{1,\hdots,|\mathcal{C}|\}$, being $n=(n_{1},\hdots,n_{|\mathcal{C}|}$), the model $\mathcal{M}$ to be trained in each client during $N_{r}$ rounds and the function $f$ to be optimized

The function \textit{minimize} applied in line 11 of Algorithm~\ref{alg:fedavgopt} optimizes the function $f$ given in Equation~\ref{eq:fx} using the Nelder-Mead method starting with the initial point $x_{0} = \boldsymbol{1}_{n}$.

%In the present work, this aggregation function has been implemented using the FL framework \textit{Flower} \cite{beutel2020flower}, so that a class inheriting from \textit{FedAvg} has been created and the optimization step for the calculation of $\alpha$ has been implemented using \textit{scipy}. 

\section{Methodology}\label{sec:methodology}

\subsection{Data under study}

The data used in this study have been obtained from an openly available database \cite{data}, so that initially we have the distribution given in Table~\ref{tab:distribution} of the data in train and test, with four categories in which the images are classified: glioma tumor, meningioma tumor, pituitary tumor and no tumor.

\begin{table}[ht]
    \centering
        \caption{Initial distribution of the data in train and test.}
    \label{tab:distribution}
    \begin{tabular}{rccc}
    \toprule
         &  \textbf{Train} & \textbf{Test} & \textbf{Total}\\
         \midrule
         \textit{Glioma tumor} & 826 & 100 & 926 \\ 
         \textit{Meningioma tumor} & 822 & 115 & 937\\ 
         \textit{Pituitary tumor} & 827 & 74 & 901\\ 
         \textit{No tumor} & 395 & 105 & 500\\ 
         \midrule
         \textit{\textbf{Total}} & 2870 & 394 & \textbf{3264} \\
    \bottomrule
    \end{tabular}
\end{table}

Note that as the data were initially centralized, we have just the train and test sets and then we have divided the whole database into 4 disjoint subsets, each of these assigned to a client. Thus, we have four randomly distributed stratified clients, which will compose the federated distributed architecture, in order to test the different aggregation functions. Subsequently, a stratified train-test split has been carried out on each client. As we aim to simulate a use case with limited data for testing the performance of the aggregation strategy proposed, we will address this during the train-test split.

\begin{figure}[ht]
    \centering
    \includegraphics[width=\linewidth]{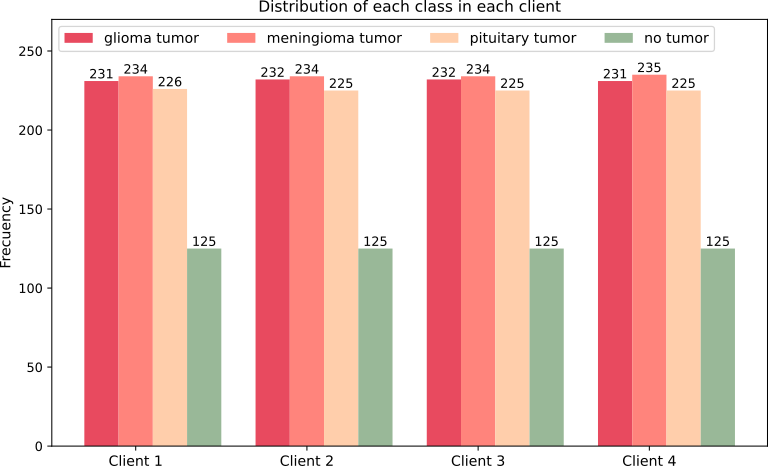}
    \caption{Distribution of the classes in the four clients.}
    \label{fig:distribution_clients}
\end{figure}

\begin{figure*}[!ht]
	\centering
    \subfigure[Axial plane.]{
		\includegraphics[width=0.47\textwidth]{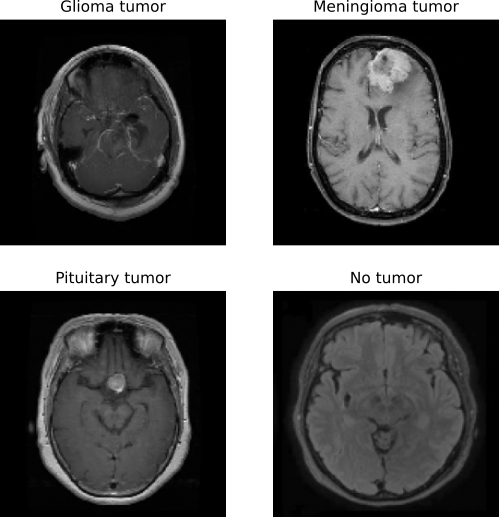}
		\label{fig:examples_client1}
	}
    \hfill
    \subfigure[Sagittal and coronal planes.]{
		\includegraphics[width=0.47\textwidth]{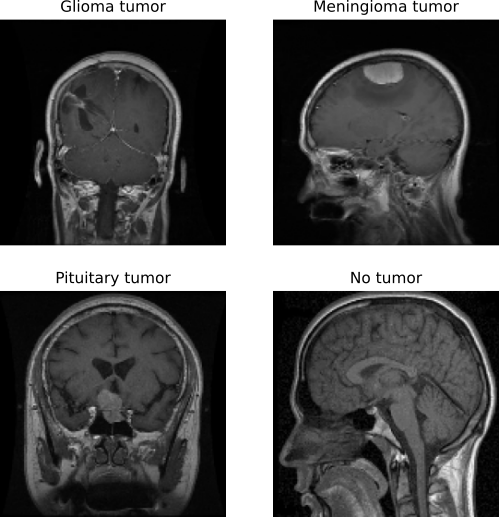}
		\label{fig:examples_client1_2}
	}
    \caption{Example of the available images for each of the four categories. Extracted from the data of the first client.}
	\label{fig:examples_client}
\end{figure*}

In view of Figure~\ref{fig:distribution_clients}, it can be seen how the distribution of the data in the different clients has been carried out in a stratified way so that the distribution of the classes in the four clients is i.i.d. These proportions will be stable when dividing into train and test conjuncts as again this has been done in a stratified way. Specifically, as we want to test the performance of the proposed strategy in case of limited data for training, we only take 20$\%$ of the data as train set (and we will test the model in the remaining 80$\%$). %Likewise, in order to validate different neural network architectures proposed according to the base model used, 20$\%$ of the train set was used as validation. 

It is important to note that in all cases images captured from different planes are available, which is important for diagnostic imaging as it provides details from different positions. For example, Figure~\ref{fig:examples_client1} shows images captured from the axial plane, while Figure~\ref{fig:examples_client1_2} also shows images captured from the sagittal plane (divides the body into left and right) and coronal plane (divides the body into anterior and posterior). In addition, it should be noted that during the pre-processing of the images they have been rescaled to a resolution of 150x150.

\subsection{Models analyzed}
In order to test the performance of the proposed strategy in different scenarios, four deep convolutional neural network architectures have been analyzed. Specifically, four base models classically used for image classification tasks have been used and a customized architecture has been created with each of these models as a base by adding the last layers to match the problem under study. The following base models have pre-trained using \textit{imagenet}. 

\begin{itemize}
    \item \textbf{VGG16}
    After applying the base model VGG16 \cite{simonyan2014very} pre-trained with image net and adapted to 150x150 resolution images with 3 color channels, we marked all layers as untrainable and then applied a flatten layer. Subsequently, in this case we applied a dense layer composed of 128 neurons and \textit{relu} activation, followed by a 0.1 rate dropout layer and by an output dense layer with 4 neurons and a \textit{softmax} activation function. As for the compilation of the model, the \textit{Adam} optimizer with learning rate 0.01, the \textit{sparse\_categorical\_crossentropy} as a loss function and the accuracy as a control metric are applied. A batch size of 16 is used at training time from the clients' side. 
    
    \textbf{Base model specifications:} Parameters: 138.4M. Depth: 16.

    \item \textbf{Inception V3}
    In this case, followed by the flatten layer applied after the base model (InceptionV3) \cite{szegedy2016rethinking}, a dense layer of 256 neurons and \textit{relu} activation is applied, followed by the output layer with 4 neurons and \textit{softmax} activation. The \textit{sparse\_categorical\_crossentropy} is again used as loss and the accuracy as metric, with an Adam optimizer in this case again with a learning rate of 0.01. Finally, a batch size of 32 is applied during the training in each client. 

    \textbf{Base model specifications:} Parameters: 23.9M. Depth: 189.

    \item \textbf{ResNet-50 V2}
    In the case of ResNet-50 V2 \cite{he2016identity}, after applying the flatten layer after the pre-trained base model, a dense layer composed of 128 sy neurons with \textit{relu} activation function is added, followed by a dropout layer of rate 0.2. The final layer is again composed of 4 neurons and \textit{softmax} activation function. The model compilation is analogous to the previous case but with a learning rate of 0.001, and applying a batch size of 16.

    \textbf{Base model specifications:} Parameters: 25.6M. Depth: 103.

    \item \textbf{DenseNet121}
    As in the previous case, once the pre-trained base model (in this case \textit{DenseNet121} \cite{huang2017densely}) is applied, a flatten layer is added, this time followed by a dense layer composed of 128 neurons with \textit{relu} activation function, followed by a dropout layer of 0.1 rate. Finally, an output dense layer with 4 neurons and \textit{softmax} activation function is applied. The model compilation is analogous to that of the ResNet-50 V2 case, also applying a batch size of 16.

    \textbf{Base model specifications:} Parameters: 8.1M. Depth: 242.

\end{itemize}

Note that different configurations have been analyzed in each case using a validation set on one of the clients and also testing the performance presented in the distributed architecture. %Thus, different final architectures have been evaluated and validated according to the base model applied, in order to customize the model that is finally shown in each case in view of the performance in the distributed case (i.e., the architecture that showed the best performance in validation was not necessarily taken, although in some cases this is the case, since it was only being evaluated for one of the clients). 

%Note that in order to implement the different architectures exposed above using the base models mentioned in each case, \textit{TensorFlow} \cite{tensorflow2015-whitepaper} (in its version 2.13 using a CPU docker image) has been used together with \textit{keras} \cite{chollet2015keras}. With \textit{keras} we can automatically download the weights for each base model once instantiated. For this purpose, the functions VGG16, InceptionV3, ResNet50V2 and DenseNet121 from \textit{tensorflow.keras} \textit{applications} have been used in order to instantiate the architecture of the base model, with an input shape of 150x150x3. The layers of the base models have been selected as non-trainable after adding the layers exposed above for each model. 

\subsection{Federated Learning configuration}

%As already explained, for performing the experiments the framework \textit{Flower} (in its version 1.8.0) will be used for orchestrating the FL architecture, while from the client side, \textit{TensorFlow} and \textit{keras} will be used for training the model. Regarding the federated learning configuration, the following setting have been established:
As already explained, for performing the experiments the framework \textit{Flower} (in its version 1.8.0) will be used for orchestrating the FL architecture. Regarding the federated learning configuration, the following setting have been established:

\begin{itemize}
    \item Number of rounds: 10.
    \item Minimum number of clients available: 4.
    \item Minimum number of clients for training: 4.
    \item Aggregated metric: \textit{accuracy}.
    \item Additional parameters:
    \begin{itemize}
        \item \textit{FedAvgM}. Server-side learning rate: 1 (default).
        
        Server-side momentum factor: 0.5. 
        \item \textit{FedOpt}: Server-side and client-side learning rate: 0.1. $\tau = 1e^{-9}$. First and second momemtum parameter: 0 (default configuration). 
        \item \textit{FedYogi}: Server-side learning rate: 0.01. Client-side learning rate: 0.0316. $\tau = 1e^{-3}$. First momemtum parameter:  0.9. Second momemtum parameter: 0.99 (default configuration).
        
        %In theses three case, the initial global model parameters introduced have been randomly initialized with a seed set to 42 for \textit{TensorFlow}, \textit{NumPy} and \textit{random}. 
    \end{itemize}
\end{itemize}

%Concerning the hardware requirements, for each base model we have created an instance in which we deploy the four clients in parallel in a distributed way (although on the same machine), as well as a server, which is exposed on port 5000. Each of the clients connects to this port to connect with the server in order to perform the federated training. All clients are located on the same machine for simplicity but it is self evident that the results can be extrapolated to the case where the clients are located on different nodes, as they need to enter the IP and the port where the server is deployed. In this case, as the data distribution has been simulated we can have them all in the same location and simulate the four clients running in parallel on the same machine. For each client a Python script has been created with a Client class inheriting from Flower client NumPyClient class, and run it using the \textit{start\_client} function from Flower. Thus, each of the 4 instances created for each base model is composed of 10 CPUs, 20GB of disk and 25GB of RAM. 

\section{Results}\label{sec:results}

\begin{table*}[!t]
    \centering
    \caption{Mean aggregated accuracy (client's test set) obtained in the 10 rounds of the FL architecture with each aggregation strategy and each base model analyzed.}
    \label{tab:acc_20train}
    %\resizebox{\linewidth}{!}{
    \begin{tabular}{ccccccc}
    \toprule
        & \multicolumn{6}{c}{\textbf{\textit{Aggregation strategy}}}\\
        \cmidrule{2-7}
        \textbf{\textit{Base model}} & \textbf{\textit{FedAvg}} & \textbf{\textit{FedAvgM}} & \textbf{\textit{FedMedian}} & \textbf{\textit{FedOpt}} & \textbf{\textit{FedYogi}} & \textbf{\textit{FedAvgOpt}}\\
        \midrule
        \textit{InceptionV3} & 0.67408 & 0.60662 & 0.67959 & 0.67263 & 0.55115 & \textbf{0.70999}\\
        \textit{VGG16} & 0.64533 & 0.76742 & 0.62734 & 0.76191 & 0.55777 & \textbf{0.77010}\\
        \textit{ResNet50V2} & 0.78622 & 0.78212 & 0.78652 & 0.78595 & 0.75992 & \textbf{0.79142} \\
        \textit{DenseNet121} & 0.79077 & 0.79100 & 0.79740 &  0.79077 & 0.76685 & \textbf{0.82209}\\
    \bottomrule
    \end{tabular}%}
\end{table*}

This section details the results obtained with each of the four base models analyzed configured as stated in Section~\ref{sec:fl}, analyzing six aggregation strategies: \textit{FedAvg}, \textit{FedAvgM}, \textit{FedMedian}, \textit{FedOpt}, \textit{FedYogi} and \textit{FedAvgOpt}. 

\begin{figure*}[!t]
	\centering
    \subfigure[InceptionV3.]{
		\includegraphics[width=0.47\textwidth]{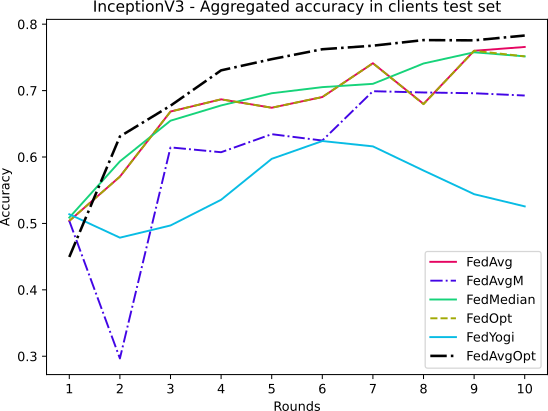}
		\label{fig:inceptionv3_mri}
	}
    \hfill
    \subfigure[VGG16.]{
		\includegraphics[width=0.47\textwidth]{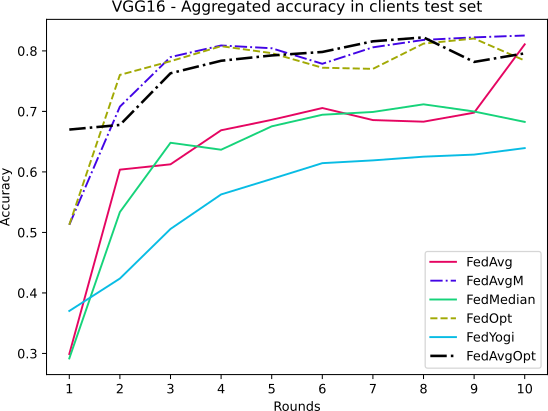}
		\label{fig:vgg16_mri}
	}
    \vfill
    \subfigure[ResNet50V2.]{
		\includegraphics[width=0.47\textwidth]{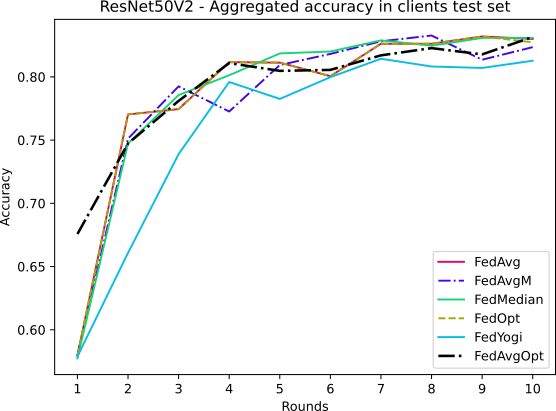}
		\label{fig:resnetv2_mri}
	}
    \hfill
    \subfigure[DenseNet121.]{
		\includegraphics[width=0.47\textwidth]{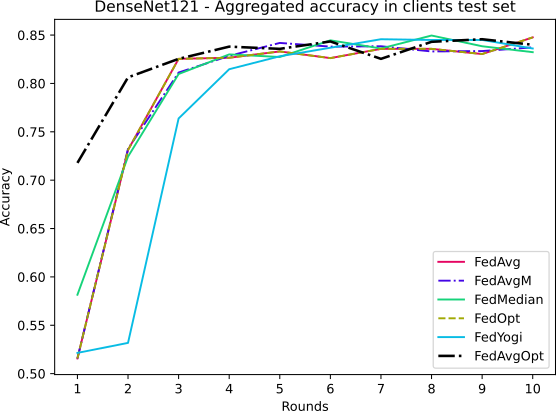}
		\label{fig:densenet121_mri}
	}
	\caption{Evolution of the aggregated accuracy in the client's test set in each round of the FL architecture with each base model.}
	\label{fig:accuracy_evolution}
\end{figure*}

The results shown in this section are the average of the aggregated accuracy obtained after each round (Figure~\ref{fig:accuracy_evolution}) as well as the average aggregated accuracy over the ten rounds (Table~\ref{tab:acc_20train}) on the test sets of each of the four clients. This aggregated accuracy has been calculated by weighting the one obtained for the test set of each client according to the number of data of each of them.

Let's start by analyzing the results with the base model InceptionV3. In view of Figure~\ref{fig:inceptionv3_mri} and Table~\ref{tab:acc_20train}, we can see that it is the model with the worst performance of the four analyzed. This may be due to a possible overfitting due to the size of the base model and the final architecture considered. We were interested in displaying these results for the convenience that seems to present the proposed aggregation approach (\textit{FedAvgOpt}) to cases where overfitting may potentially occur. Especially it is interesting to compare the performance with the aggregated accuracy obtained with \textit{FedAvgM} and \textit{FedYogi}, which despite receiving some initial global parameters for the model, do not achieve stability over the rounds nor convergence of the model as if it is clearly seen to occur in the curve of \textit{FedAvgOpt} in Figure~\ref{fig:inceptionv3_mri}. On average, with this base model \textit{FedAvgOpt} is at least 3$\%$ more accurate than the other strategies considered, rising to mote than 10$\%$ and 15$\%$ for \textit{FedAvgM} and \textit{FedYogi} respectively.

Next, the case of VGG16 stands out, where it can be observed that the simplest aggregation strategies do not achieve good convergence (see Figure~\ref{fig:vgg16_mri}). This is clearly seen with \textit{FedAvg} and \textit{FedMedian}, while those functions that receive global parameters in the input (\textit{FedAvgM} and \textit{FedOpt}) achieve a better generalization capability and a faster convergence, except for \textit{FedYogi}, which again as with \textit{InceptionV3}, fails to converge. \textit{FedAvgOpt}, like \textit{FedAvgM} and \textit{FedOpt}, achieves good convergence and predictive capability, although it does not need such global parameters to achieve it. Thus, on average for the 10 rounds, the latter is the strategy that provides the best mean accuracy, being also the one that starts from a higher accuracy after the first round, exceeding that of the rest of the strategies in that round by more than 15$\%$. 

The first thing we can highlight from Table~\ref{tab:acc_20train} is that on average for the 10 rounds, the aggregated results for test with \textit{FedAvgOpt} is better than with the other five strategies classically used in the literature. This is fulfilled for the four base models and the architectures proposed for each of them, with a significant improvement for InceptionV3, where none of the classical aggregation functions reaches an accuracy of 68$\%$ and \textit{FedAvgOpt} exceeds 70$\%$. Furthermore, with DenseNet121 and \textit{FedAvgOpt} we reach an accuracy greater than 82$\%$ while the rest of aggregation functions do not reach 80$\%$. Finally, the average accuracy with \textit{FedAvgOot} is higher than 77$\%$ and 79$\%$ respectively for \textit{VGG16} and \textit{ResNet50V2}, while using the other aggregation strategies these values are not reached in any of the five approaches

In view of Figure~\ref{fig:accuracy_evolution}, it can be seen that in the case of \textit{VGG16}, \textit{DenseNet121} and \textit{ResNet50V2}, the accuracy in the first round is much higher with \textit{FedAvgOpt} than with the rest of the strategies, being more than 10$\%$ higher. With \textit{InceptionV3}, only in the first round the accuracy is slightly lower than with the other approaches, but the convergence is faster and higher during the following rounds. Note that the \textit{FedAvgOpt} curves in Figures~\ref{fig:inceptionv3_mri}, \ref{fig:vgg16_mri}, \ref{fig:resnetv2_mri} and \ref{fig:densenet121_mri} are more stable and regular, with fewer significant peaks or troughs. For example, with \textit{InceptionV3} there are significant drops in \textit{FedAvg}, \textit{FedAvgM}, \textit{FedOpt} and \textit{FedYogi}, while with \textit{FedAvgOpt} there is a good and stable convergence.

Thus, in view of the results obtained with the different base models and aggregation functions, we can conclude that in this case the aggregation with \textit{FedAvgOpt} is a very significant option to be taken into account to carry out the aggregation of the models in the federated architecture. Specifically, this has been clearly proven in this case where the distribution of the clients is i.i.d. and it can be seen both in the mean accuracy over the 10 rounds analyzed (see Table~\ref{tab:acc_20train}) and in the evolution during the rounds (see Figure~\ref{fig:accuracy_evolution}). The proper performance of \textit{FedAvgOpt} has been tested both in cases of models composed of a high number of parameters (VGG16 with 138.5M of parameters), as well as in cases with a smaller number of parameters (\textit{DenseNet121} with 8.1M of parameters), and in deeper (\textit{DenseNet121} with depth 242, \textit{InceptionV3} 189, \textit{ResNet50V2} with 103) or less deep models (\textit{VGG16} with depth 16).

\section{Conclusions and future work}\label{sec:conclusions}
 
In this work we have presented a new aggregation strategy that seeks to enhance the convergence of classic functions used in the literature, showing very promising results for the case of brain MRI analyzed. Overall, we can conclude that in the use case studied \textit{FedAvgOpt} is the most stable aggregation strategy of the 6 analyzed across the 4 base models reviewed. For example, \textit{FedAvgM} is very close to its performance in the cases of VGG16 and ResNet-50 V2, but it has a very unstable performance with Inception V3. The only strategy that has a stable performance throughout the 4 models similar to that achieved with \textit{FedAvgOpt} is \textit{FedMedian}, but it has a very low accuracy in the case of VGG16, and takes longer than \textit{FedAvgOpt} to stabilize (just refer to the performance in round 1 of ResNet and DenseNet121, as well as the curve for Inception V3, where the accuracy is always lower than that obtained with \textit{FedAvgOpt} from round 2 onwards). On the other hand, \textit{FedYogi} has a very poor performance in all four cases, being only relatively stable in ResNet-50 V2 and DenseNet121, but with a lower performance than the rest of the analyzed strategies in all 4 cases.

As highlighted in Section~\ref{sec:sota}, it has been chosen to analyze five aggregation strategies that do not require client-side modifications in their implementation. As reviewed in the state of the art, there are other techniques that can be considered, such as \textit{Scaffold}, \textit{Moon}, or \textit{FedAdam} and \textit{FedAdagrad}. %In particular, we began by analyzing \textit{FedAdam} for the case of the best accuracy obtained in the tests performed, using DenseNet121, however, since the accuracy did not end up converging for this particular example with these strategies, it was decided not to add it to this comparative study. 

As future work, we will continue with the application of \textit{FedAvgOpt} to other use cases, especially to analyze its performance on non i.i.d. settings. It is also planned to extend the study to more base models and other aggregation functions, although in relation to the former, four widely different models in terms of number of parameters and depth have been tested. Finally, it is of interest to take this research to a real use case to allow collaboration between different medical centers or hospitals in model building without data sharing, resulting in a tool that can support health professionals in their diagnoses, with the models been trained respecting the privacy of the corresponding patients involved, as data will not leave each data center. \\

\section*{Acknowledgment}

\noindent We would like to thank the funding and support from the AI4EOSC project (funded by the European Union's Horizon Europe research and innovation programme under grant agreement number 101058593) and from the EOSC-SIESTA project (funded by the European Union (Horizon Europe) under grant agreement number 101131957).

\bibliography{preprint}

\end{document}